\pdfoutput=1

\documentclass[11pt,a4paper]{article}
\usepackage{emnlp2020}
\usepackage[ruled,vlined]{algorithm2e}
\usepackage{url}
\usepackage{times}
\usepackage{latexsym}

\usepackage{graphicx}
\usepackage{microtype}

\aclfinalcopy 

\graphicspath{{figures/}}

\title{{NeuralQA}: A Usable Library for Question Answering (Contextual Query Expansion + BERT) on Large Datasets}

\author{ 
  Victor Dibia \\
  Cloudera Fast Forward Labs \\ 
  \texttt{vdibia@cloudera.com} \\}

\date{}

\begin{document}

\maketitle
\begin{abstract}



Existing tools for Question Answering (QA) have challenges that limit their use in practice. They can be complex to set up or integrate with existing infrastructure, do not offer configurable interactive interfaces, and do not cover the full set of subtasks that frequently comprise the QA pipeline (query expansion, retrieval, reading, and explanation/sensemaking). To help address these issues, we introduce NeuralQA - a usable library for QA on large datasets. NeuralQA integrates well with existing infrastructure (e.g.,  ElasticSearch instances and reader models trained with the HuggingFace Transformers API) and offers helpful defaults for QA subtasks. It introduces and implements contextual query expansion (CQE) using a masked language model (MLM) as well as relevant snippets (\(RelSnip\)) - a method for condensing large documents into smaller passages that can be speedily processed by a document reader model. Finally, it offers a flexible user interface to support workflows for research explorations (e.g., visualization of gradient-based explanations to support qualitative inspection of model behaviour) and large scale search deployment.  
Code and documentation for NeuralQA is available as open source on  \href{https://github.com/victordibia/neuralqa}{Github}. 




\end{abstract}

\section{Introduction}


\begin{figure}[t!]
  \centering
  \includegraphics[width=\columnwidth]{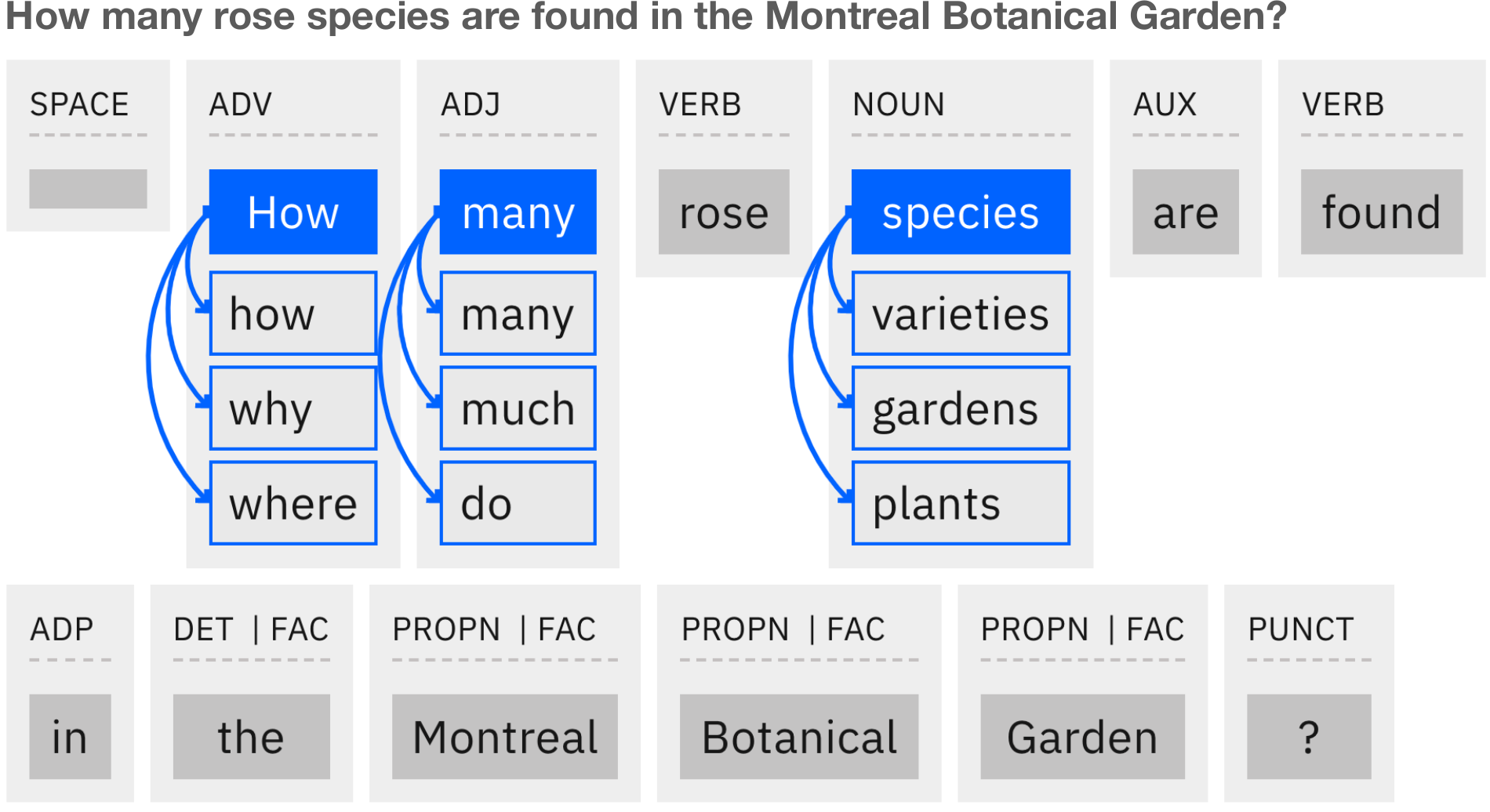}
  \caption{NeuralQA implements Contextual Query Expansion (CQE \ref{sec:contextualqueryexpansion}) using Masked Language Models (MLM)  and offers a visualization to explain behaviour. A rule set is used to determine which tokens are \textit{candidates} for expansion (solid blue box); each \textit{candidate} is iteratively masked, and an MLM is used to identify expansion terms (blue outline box).
  \label{fig:expander}} 
\end{figure}

The capability of providing \textit{exact} answers to queries framed as natural language questions can significantly improve the user experience in many real world applications. Rather than sifting through lists of retrieved documents, automatic QA (also known as reading comprehension) systems can surface an exact answer to a query, thus reducing the cognitive burden associated with the standard search task. This capability is applicable in extending conventional information retrieval systems (search engines) and also for emergent use cases, such as open domain conversational AI systems \cite{gao2018neural, qu2019bert}. For enterprises, QA systems that are both fast and precise can help unlock knowledge value in large unstructured document collections.  
Conventional methods for open domain QA \cite{yang2015wikiqa, yang2019end} follow a two-stage implementation - (i) a \textbf{retriever} that returns a subset of relevant documents. Retrieval is typically based on sparse vector space models such as BM25 \cite{robertson2009probabilistic} and TF-IDF \cite{chen2017reading}; (ii) a machine reading comprehension model (\textbf{reader}) that identifies spans from each document which contain the answer. While sparse representations are fast to compute, they rely on exact keyword match, and suffer from the \textit{vocabulary mismatch} problem - scenarios where the vocabulary used to express a query is different from the vocabulary used to express the same concepts within the documents. 
To address these issues, recent studies have proposed neural ranking \cite{lee-etal-2018-ranking, kratzwald-etal-2019-rankqa} and retrieval methods \cite{karpukhin2020dense, lee2019latent, guu2020realm}, which rely on dense representations.  

However, while dense representations show significantly improved results, they introduce additional complexity and latency, which limits their practical application. For example, \citet{guu2020realm} require a specialized MLM pretraining regime, as well as a supervised fine-tuning step, to obtain representations used in a retriever. Similarly \citet{karpukhin2020dense} use dual encoders in learning a dense representation for queries and all documents in the target corpus. Each of these methods require additional infrastructure to compute dense representation vectors for all documents in the target corpus as well as implement efficient similarity search at run time. In addition, transformer-based architectures \cite{vaswani2017attention} used for dense representations are unable to process long sequences due to their self-attention operations which scale quadratically with sequence length. As a result, these models require that documents are indexed/stored in small paragraphs. For many use cases, meeting these requirements (rebuilding retriever indexes, training models to learn corpus specific representations, precomputing representations for all indexed documents) can be cost-intensive. These costs are hard to justify, given that simpler methods can yield comparable results \cite{lin2019neural, WeissenbornWS17}. Furthermore, as reader models are applied to domain-specific documents, they fail in counter-intuitive ways. It is thus valuable to offer visual interfaces that support debugging or sensemaking of results (e.g., explanations for \textit{why} a set of documents were retrieved or \textit{why} an answer span was selected from a document). While several libraries exist to explain NLP models, they do not integrate interfaces that help users make sense of both the query expansion, retriever and the reader tasks. Collectively, these challenges can hamper experimentation with QA systems and the integration of QA models into practitioner workflows.

In this work, we introduce NeuralQA to help address these limitations. Our contributions are summarized as follows:

\begin{itemize}
\itemsep0em 
    \vspace{-0.1cm}\item An easy to use, end-to-end library for implementing QA systems. It integrates methods for query expansion, document retrieval (ElasticSearch\footnote{ElasticSearch {https://www.elastic.co}}), and document reading (QA models trained using the HuggingFace Transformers API \cite{Wolf2019HuggingFacesTS}). It also offers an interactive user interface for sensemaking of results (retriever + reader). NeuralQA is \href{https://github.com/victordibia/neuralqa}{open source  and released under the MIT License}. 
   \vspace{-0.1cm} \item To address the vocabulary mismatch problem, NeuralQA introduces and implements a method for contextual query expansion (CQE), using a masked language model (MLM) fine-tuned on the target document corpus (see Fig  \ref{fig:expander}). Early qualitative results show CQE can surface relevant additional query terms that help improve recall and require minimal changes for integration with existing retrieval infrastructure.
    \vspace{-0.1cm}\item In addition, we implement \(RelSnip\), a simple method for extracting relevant snippets from retrieved passages before feeding it into a document reader. This, in turn, reduces the latency required to chunk and read lengthy documents. Importantly, these options offer the opportunity to improve latency and recall, with no changes to existing retriever infrastructure.     
    


\end{itemize}

Overall, NeuralQA complements a line of end-to-end applications that improve QA system deployment \cite{akkalyoncu-yilmaz-etal-2019-applying,yang2019end} and provide visual interfaces for understanding machine learning models \cite{Wallace2019AllenNLP, seq2seqvisv1,madsen2019visualizing, Dibia2020Anomagram, Dibia2020Convnetplayground}.



\section{The Question Answering Pipeline}

\label{relatedwork} 
There are several subtasks that frequently comprise the QA pipeline and are implemented in NeuralQA.  


\subsection{Document Retrieval}

The first stage in the QA process focuses on retrieving a list of candidate passages, which are subsequently processed by a reader. Conventional approaches to QA apply representations from sparse vector space models (e.g., BM25, TF-IDF) in identifying the most relevant document candidates. For example, \citet{chen2017reading} introduce an end-to-end system combining TF-IDF retrieval with a multi-layer RNN for document reading. This is further improved upon by \citet{yang2019end}, who utilize BM25 for retrieval with a modern BERT transformer reader. However, sparse representations are keyword dependent, and suffer from the \textit{vocabulary mismatch} problem in information retrieval (IR); given a query \(Q\) and a relevant document \(D\), a sparse retrieval method may fail to retrieve \(D\) if \(D\) uses a different vocabulary to refer to the same content in \(Q\). Furthermore, given that QA queries are under-specified by definition (users are searching for unknown information), sparse representations may lack the contextual information needed to retrieve the most relevant documents. To address these issues, a set of related work has focused on methods for re-ranking retrieved documents to improve recall \cite{wang2018opendomain,kratzwald-etal-2019-rankqa}. More recently, there have been efforts to learn representations of queries and documents useful for retrieval. \citet{lee2019latent} introduce an inverse cloze task for pretraining encoders used to create static embeddings that are indexed and used for similarity retrieval during inference. Their work is further expanded by \citet{guu2020realm} who introduce non-static representations that are learned simultaneous to reader fine-tuning. Finally, \citet{karpukhin2020dense} use dual encoders for retrieval: one encoder that learns to map queries to a fixed dimension vector, and another that learns to map documents to a similar fixed-dimension vector (such that representations for similar query and documents are close).
 
 
\subsection{Query Expansion}
In addition to re-ranking and dense representation retrieval, query expansion methods have also been proposed to help address the vocabulary mismatch problem. They serve to identify additional relevant query terms, using a variety of sources - such as the target corpus, external dictionaries (e.g., WordNet), or historical queries.
Existing research has explored how implicit information contained in queries can be leveraged in query expansion. For example, \citet{lavrenko2017relevance, lv2010positional} show how a relevance model (RM3) can be applied for query expansion and improve retrieval performance. More recently, \cite{lin2019neural} also show that the use of a well-tuned relevance model such as RM3 \cite{lavrenko2017relevance, abdul2004umass} results in performance at par with complex neural retrieval methods.
Word embeddings have been explored as a potential method for query expansion, as well. In their work, \citet{kuzi2016query} train a word2vec \cite{mikolov2013distributed} CBOW model on their search corpora and use embeddings to identify expansion terms that are either semantically related to the query as a whole or to its terms. Their results suggest that a combination of word2vec embeddings and a relevance model (RM3) provide good results. However, while word embeddings trained on a target corpus are useful, they are static and do not take into consideration the context of the words in a specific query. In this work, we propose an extension to this direction of thought and explore how contextual embeddings produced by an MLM, such as BERT \cite{devlin2018bert}, can be applied in generating query expansion terms.


\subsubsection{Document Reading}
Recent advances in pretrained neural language models, like BERT \cite{vaswani2017attention} and GPT \cite{radford2019language}, have enabled robust contextualized representation of natural language, which, in turn, have enabled significant performance increases on the QA task. Each QA model (reader) consists of a base representation and an output feedforward layer which produces two sets of scores: (i) scores for each input token that indicate the likelihood of an answer span starting at the token offset, and (ii) scores for each input token that indicate the likelihood of an answer span ending at the token offset.

\section{NeuralQA System Architecture}

\begin{figure*}[ht]
  \centering
  \includegraphics[width=\textwidth]{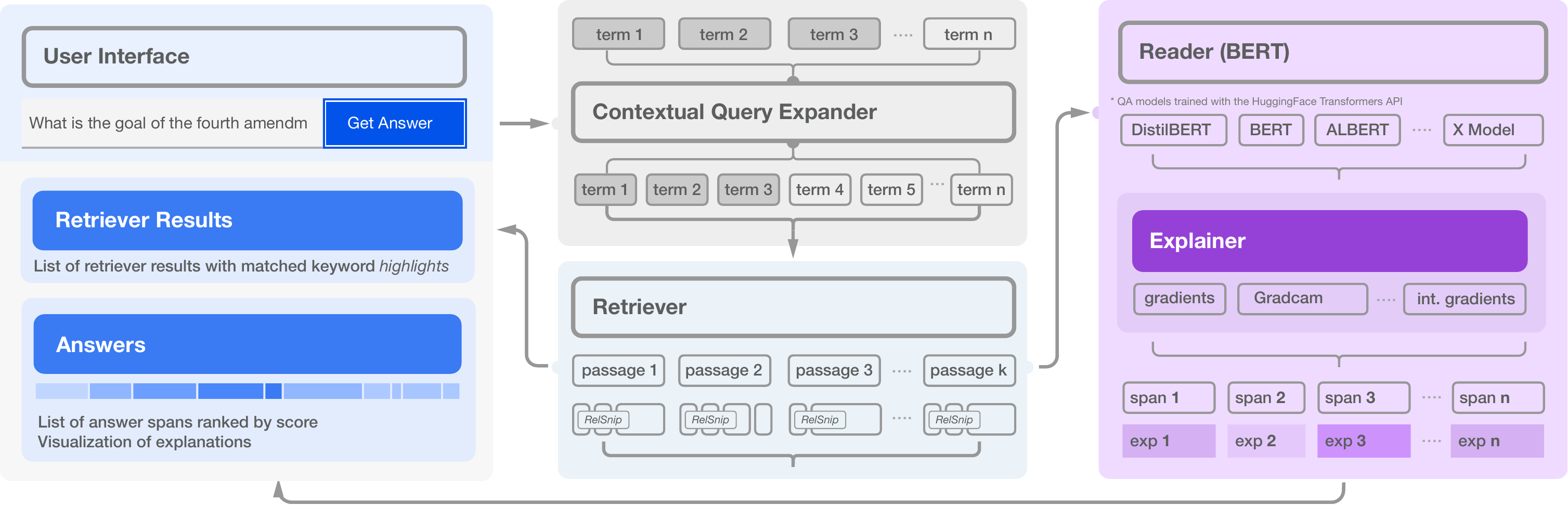}
  \caption{The NeuralQA Architecture is comprised of four primary modules. (a) User interface: enables user queries and visualizes results from the retriever and reader (b) Contextual Query Expander: offers options for generating query expansion terms using an MLM (c) Retriever: leverages the BM25 scoring algorithm in retrieving a list of candidate passages; it also optionally condenses lengthy documents to smaller passages via \ref{sec:relsnip} \(RelSnip\). (d) Document Reader: identifies answer spans within documents (where available) and provides explanations for each prediction.
  \label{fig:architecture}} 
\end{figure*}

In this section, we review the architecture for NeuralQA, as well as design decisions and supported workflows. The core modules for NeuralQA (Fig. \ref{fig:architecture}) include a user interface, retriever, expander, and reader. Each of these modules are implemented as extensible python classes (to facilitate code reuse and incremental development), and are exposed as REST API endpoints that can be either consumed by 3rd party applications or interacted with via the NeuralQA user interface.

\subsection{Retriever}
The retriever supports the execution of queries on an existing ElasticSearch instance, using the industry standard BM25 scoring algorithm. 

\label{sec:relsnip}
\subsubsection{Condensing Passages with \(RelSnip\)}
In practice, open corpus documents can be of arbitrary length (sometimes including thousands of tokens) and are frequently indexed for retrieval \textit{as is}. On the other hand, document reader models have limits on the maximum number of tokens they can process in a single pass (e.g., BERT-based models can process a maximum of 512 tokens). Thus, retrieving large documents can incur latency costs, as a reader will have to first split the document into manageable \textit{chunks}, and then process each \textit{chunk} individually. To address this issue, NeuralQA introduces \(RelSnip\), a method for constructing smaller documents from lengthy documents. \(RelSnip\) is implemented as follows: For each retrieved document, we apply a highlighter (\href{ https://lucene.apache.org/core/7_3_1/highlighter/org/apache/lucene/search/uhighlight/UnifiedHighlighter.html}{Lucene Unified Highlighter}), which breaks the document into fragments of size \(k_{frag}\) and uses the BM25 algorithm to score each fragment as if they were individual documents in the corpus. Next, we concatenate the top \(n\) fragments as a new document, which is then processed by the reader. \(RelSnip\) relies on the simplifying assumption that fragments with higher match scores contain more relevant information. As an illustrative example, \(RelSnip\) can yield a document of 400 tokens (depending on \(k_{frag}\) and \(n\) ) from a document containing 10,000 tokens. In practice, this can translate to ~25x increase in speed.

\subsection{Expander}
\subsubsection{Contextual Query Expansion (CQE)}
\label{sec:contextualqueryexpansion}
CQE relies on the assumption that an MLM which has been fine-tuned on the target document corpus contains implicit information \cite{petroni-etal-2019-language} on the target corpus. The goal is to exploit this information in identifying relevant query expansion terms. Ideally, we want to expand a query, such that expansion tokens serve to increase recall, while adding minimal noise and without significantly altering the semantics of the original query. We implement CQE as follows: First, we identify a set of expansion candidate tokens. For each token \(t_i\) in the query \(t_{query}\), we use the SpaCy \cite{spacy2} library to infer its part of speech tag \(t_{i_{pos}}\) and apply a filter \(f_{rule}\)  to determine if it is added to a list of candidate tokens for expansion \(t_{candidates}\).  Next, we construct intermediate versions of the original query, in which each token in \(t_{candidates}\) is masked, and an MLM (BERT) predicts the top \(n\) tokens that are contextually most likely to complete the query. These predicted tokens \(t_{expansion}\) can then be added to the original query as expansion terms.

\begin{figure}[t!]
  \centering
  \includegraphics[width=\columnwidth]{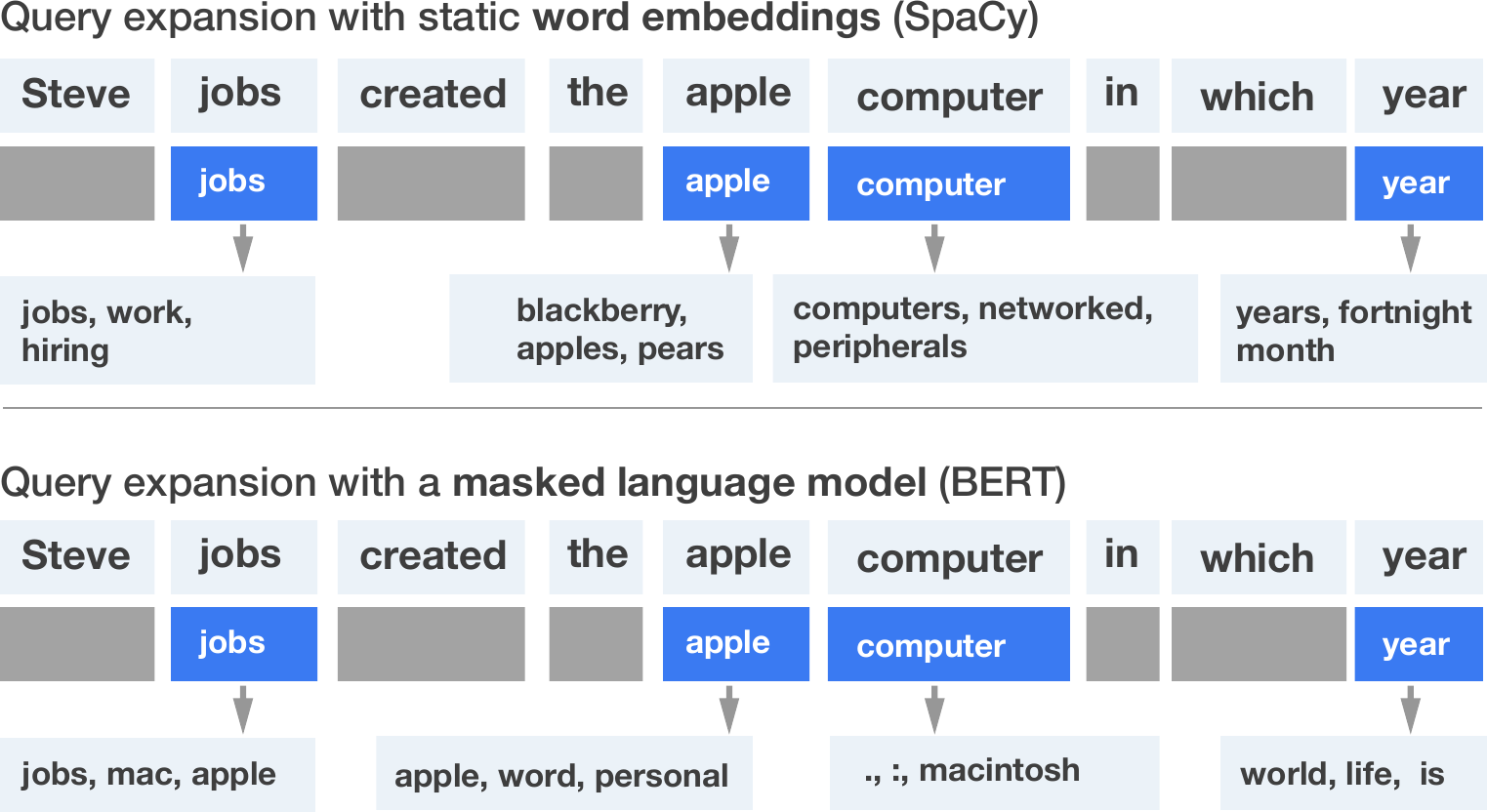}
  \caption{Examples of qualitative results from applying query expansion: (a) Query expansion using SpaCy word embeddings to identify the most similar words for each expansion candidate token. This approach yields terms with low relevance (e.g., terms related to work (jobs, hiring) and fruits (apple, blackberry, pears) are not relevant to the current query context) (b) Query expansion using an MLM (BERT). This approach yields terms that are absent in the original query (e.g., mac, macintosh, personal) but are, \textit{in general}, relevant to the current query.
  \label{fig:mlm}} 
\end{figure}

To minimize the chance of introducing spurious terms that are unrelated to the original query, we find that two quality control measures are useful. First, we leverage confidence scores returned by the MLM and only accept expansion tokens above a certain threshold (e.g., \(k_{thresh} =0.5\)) where \(k_{thresh}\)  is a hyperparameter. Secondly, we find that a conservative filter in selecting token expansion candidates can mitigate the introduction of spurious terms. Our filter rule \(f_{rule}\) currently only expands tokens that are either nouns or adjectives \(t_{i_{pos}} \in ({noun, adj}) \) and are not named entities; tokens for other parts of speech are not expanded. Finally, the list of expansion terms are further cleaned by the removal of duplicate terms, punctuation, and stop words. Fig. \ref{fig:mlm} shows a qualitative comparison of query expansion terms suggested by a static word embedding and an MLM for a given query. The NeuralQA interface offers a user-in-the-loop visualization of CQE which highlights POS tags for each token to help the user make sense of expansion values. The user can then select expansion candidates for inclusion in retrieval.

\subsection{Reader}
The reader module implements an interface for predicting answer spans, given a query and context documents. Underneath, it loads any QA model trained using the HuggingFace Transformers API \cite{Wolf2019HuggingFacesTS}. Documents that exceed the maximum token size for the reader are automatically split into chunks with a configurable stride and answer spans provided for each chunk. All answers are then sorted, based on an associated score (start and end token softmax probabilities). Finally, each reader model provides a method that generates gradient-based explanations (Vanilla Gradients \cite{simonyan2013deep,erhan2009visualizing,baehrens2010explain}). 

\begin{figure*}[ht] 
  \includegraphics[width=\textwidth]{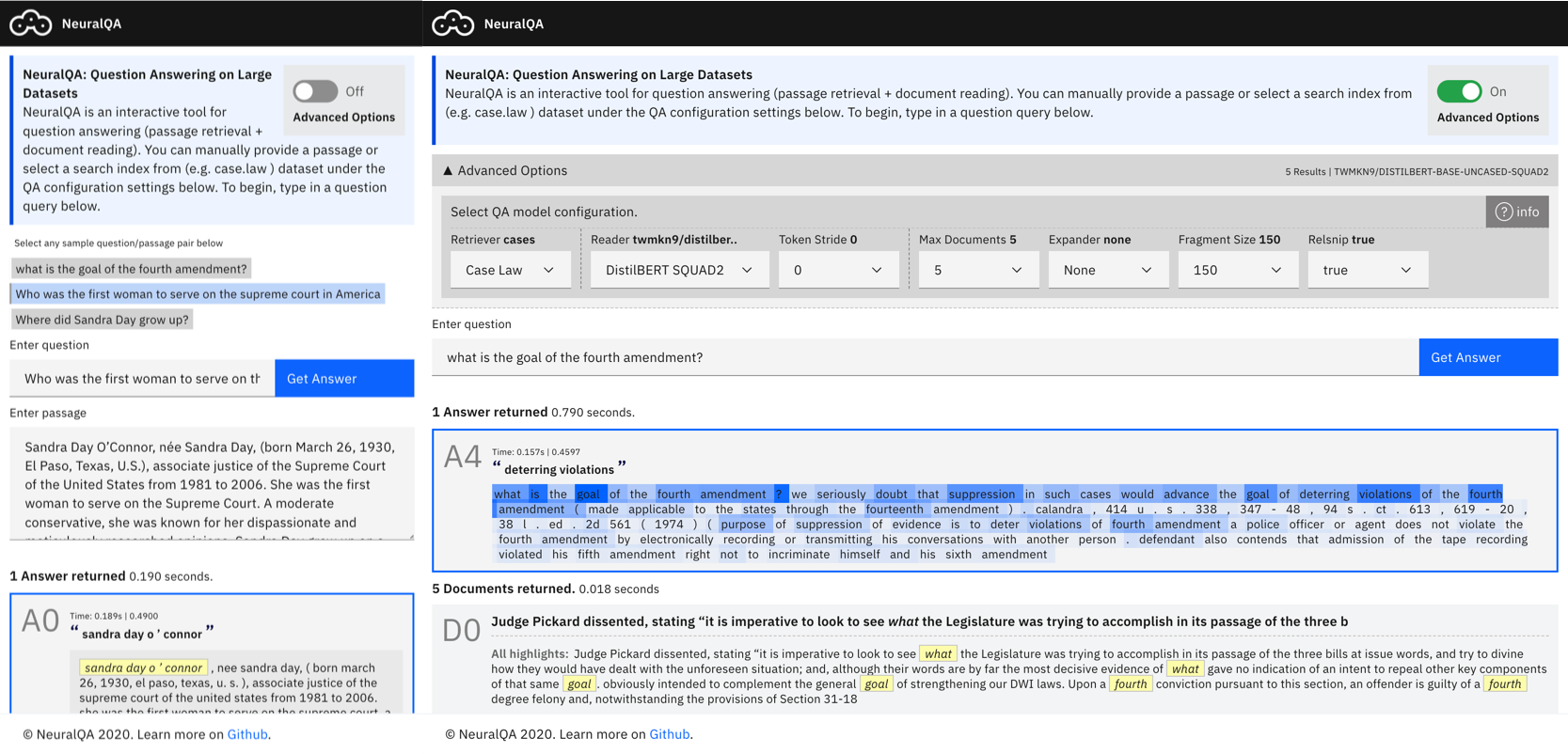}
  \caption{ 
  The NeuralQA UI. a.) Basic view (mobile) for closed domain QA, i.e., the user provides a question \textit{and} passage. b.) Advanced options view (desktop mode) for open domain QA. The user can select the retriever (e.g., \# of returned documents, toggle \(RelSnip\), fragment size \(k_{frag}\)), set expander and reader parameters (BERT reader model, token stride size)). View also shows a list of returned documents (D0-D4) with highlights that match query terms; a list of answers (A0) with  gradient-based explanation of which tokens impact the selected answer span. 
  \label{fig:uiscreen}}
\end{figure*}

\subsubsection{User Interface }
The NeuralQA user interface (Fig. \ref{fig:uiscreen}) seeks to aid the user in performing queries and in sensemaking of underlying model behaviour. As a first step, we provide a visualization of retrieved document highlights that indicate what portions of the retrieved document contributed to their relevance ranking. Next, following work done in AllenNLP Interpret\cite{Wallace2019AllenNLP}, we implement gradient-based explanations that help the user understand what sections of the input (question and passage) were most relevant to the choice of answer span. We do not use attention weights, as they have have been shown to be unfaithful explanations of model behaviour \cite{jain2019attention, serrano-smith-2019-attention} and not intuitive for end user sensemaking. We also implement a document and answer tagging scheme that indicates the source document from which an answer span is derived.


NeuralQA is scalable, as it is built on industry standard OSS tools that are designed for scale (\href{elastic.co}{ElasticSearch}, \href{https://github.com/huggingface/transformers}{HuggingFace Transformers API}, \href{https://fastapi.tiangolo.com/}{FastAPI}, \href{https://www.uvicorn.org/}{Uvicorn} asgi web server). We have tested deployments of NeuralQA on docker containers running on CPU machine clusters which rely on ElasticSearch clusters. The UI is responsive and optimized to work on desktop, as well as on mobile devices.

 \subsection{Configuration and Workflow}

NeuralQA implements a command line interface for instantiating the library, and a declarative approach for specifying the parameters for each module. At run time, users can provide a command line argument specifying the location of a configuration YAML file\footnote{ A sample configuration file can be found on \href{https://github.com/victordibia/neuralqa/blob/master/neuralqa/config_default.yaml}{Github.}}. If no configuration file is found in the provided location and in the current folder, NeuralQA will create a default configuration file that can be subsequently modified. As an illustrative example, users can configure properties of the user interface (views to show or hide, title and description of the page, etc.), retriever properties (a list of supported retriever indices), and reader properties (a list of supported models that are loaded into memory on application startup).

\subsubsection{User Personas}
NeuralQA is designed to support use cases and personas at various levels of complexities. We discuss two specific personas briefly below.
\\\textbf{Data Scientists}: Janice, a data scientist, has extensive experience applying a collection of machine learning models to financial data. Recently, she has started a new project, in which the deliverable includes a QA model that is skillful at answering factoid questions on financial data. As part of this work, Janice has successfully fine-tuned a set of transformer models on the QA task, but would like to better understand how the model behaves. More importantly, she would like to enable visual interaction with the model for her broader team. To achieve this, Janice hosts NeuralQA on an internal server accessible to her team. Via a configuration file, she can specify a set of trained models, as well as enable user selection of reader/retriever parameters. This workflow also extends to other user types (such as hobbyists, entry level data scientists, or researchers) who want an interface for qualitative inspection of custom reader models on custom document indices.  
\\\textbf{Engineering Teams}: Candice manages the internal knowledge base service for her startup. They have an internal ElasticSearch instance for search, but would like to provide additional value via QA capabilities. To achieve this, Candice provisions a set of docker containers running instances for NeuralQA and then modifies the frontend of their current search application to make requests to the NeuralQA REST API and serve answer spans.
 




\subsection{Related Work}
QA systems that integrate deep learning models remain an active area of research and practice. For example, AllenNLP Interpret \cite{Wallace2019AllenNLP} provides a demonstration interface and sample code for interpreting a set of AllenNLP models across multiple tasks. Similarly, \citet{chakravarti-etal-2019-cfo} provide a gRPC-based orchestration flow for QA. However, while these projects provide a graphical user interface (GUI), their installation process is complex and requires specialized code to adapt them to existing infrastructure, such as retriever instances. Several open source projects also offer a programmatic interface for inference (e.g.,   \href{https://huggingface.co/transformers/main_classes/pipelines.html}{ HugginFace Pipelines}), as well as joint retrieval paired with reading (e.g.,   \href{https://github.com/deepset-ai/haystack/}{ Deepset Haystack}). 
NeuralQA makes progress along these lines, by providing an extensible code base, a low-code declarative configuration interface, tools for query expansion and a visual interface for sensemaking of results. It supports a local research/development workflow (via the \href{https://pypi.org/project/neuralqa/}{pip}) package manager  and scaled deployment via containerization (we provide a Dockerfile). We believe this ease of use can serve to remove barriers to experimentation for researchers, and accelerate the deployment of QA interfaces for experienced teams.



\section{Conclusion}

In this paper, we presented NeuralQA - a usable library for question answering on large datasets. 
NeuralQA is useful for developers interested in qualitatively exploring QA models for their custom datasets, as well as for enterprise teams seeking a flexible QA interface/API for their customers. 
NeuralQA is under active development, and roadmap features include support for a \href{https://lucene.apache.org/solr/}{Solr} retriever, additional model explanation methods and additional query expansion methods such as RM3 \cite{lavrenko2017relevance}. Future work will also explore empirical evaluation of our CQE and \(RelSnip\) implementation to better understand their strengths and limitations.








\section*{Acknowledgments}

The author thanks Melanie Beck, Andrew Reed, Chris Wallace, Grant Custer, Danielle Thorpe and other members of the Cloudera Fast Forward team for their valuable feedback.

\bibliography{paper}
\bibliographystyle{acl_natbib}

\end{document}